\documentclass[runningheads,a4paper]{llncs}

\usepackage{amssymb}
\usepackage{graphicx}
\usepackage{amsmath}
\usepackage{caption} 
\usepackage{subcaption}
\usepackage[colorinlistoftodos]{todonotes}
\usepackage{array}
\newcolumntype{M}[1]{>{\raggedright\arraybackslash}m{#1}}

\usepackage{url}
\urldef{\mailsa}\path|{alfred.hofmann, ursula.barth, ingrid.haas, frank.holzwarth,|
\urldef{\mailsb}\path|anna.kramer, leonie.kunz, christine.reiss, nicole.sator,|
\urldef{\mailsc}\path|erika.siebert-cole, peter.strasser, lncs}@springer.com|

\begin{document}

\mainmatter  

\title{Fast Multiple Landmark Localisation Using a Patch-based Iterative Network}

\titlerunning{Fast Multiple Landmark Localisation Using PIN}

%
%
\author{Yuanwei Li\inst{1}\and Amir Alansary\inst{1}\and Juan J. Cerrolaza\inst{1}\and Bishesh Khanal\inst{2}\and\\
Matthew Sinclair\inst{1}\and Jacqueline Matthew\inst{2}\and Chandni Gupta\inst{2}\and Caroline Knight\inst{2}\and Bernhard Kainz\inst{1}\and Daniel Rueckert\inst{1}}
\authorrunning{Y. Li et al.}

\institute{Biomedical Image Analysis Group, Imperial College London, UK
\and
School of Biomedical Engineering \& Imaging Sciences, King's College London, UK
}

%
%

\toctitle{Lecture Notes in Computer Science}
\tocauthor{Authors' Instructions}
\maketitle

\begin{abstract}
We propose a new Patch-based Iterative Network (PIN) for fast and accurate landmark localisation in 3D medical volumes. PIN utilises a Convolutional Neural Network (CNN) to learn the spatial relationship between an image patch and anatomical landmark positions. During inference, patches are repeatedly passed to the CNN until the estimated landmark position converges to the true landmark location. PIN is computationally efficient since the inference stage only selectively samples a small number of patches in an iterative fashion rather than a dense sampling at every location in the volume. Our approach adopts a multi-task learning framework that combines regression and classification to improve localisation accuracy. We extend PIN to localise multiple landmarks by using principal component analysis, which models the global anatomical relationships between landmarks. We have evaluated PIN using 72 3D ultrasound images from fetal screening examinations. PIN achieves quantitatively an average landmark localisation error of 5.59mm and a runtime of 0.44s to predict 10 landmarks per volume. Qualitatively, anatomical 2D standard scan planes derived from the predicted landmark locations are visually similar to the clinical ground truth. Source code is publicly available at \url{https://github.com/yuanwei1989/landmark-detection}.
\end{abstract}

\section{Introduction}
Anatomical landmark localisation is a key challenge for many medical image analysis tasks. Accurate landmark identification can be used for (a) extracting biometric measurements of anatomical structures, (b) landmark-based registration of 3D volumes, (c) extracting 2D clinical standard planes from 3D volumes and (d) initialisation of tasks such as image segmentation. However, manual landmark detection is time-consuming and suffers from high observer variability. Thus, there is a need to develop automatic methods for fast and accurate landmark localisation. \begin{figure}[htb]
\centering
\includegraphics[width=\linewidth]{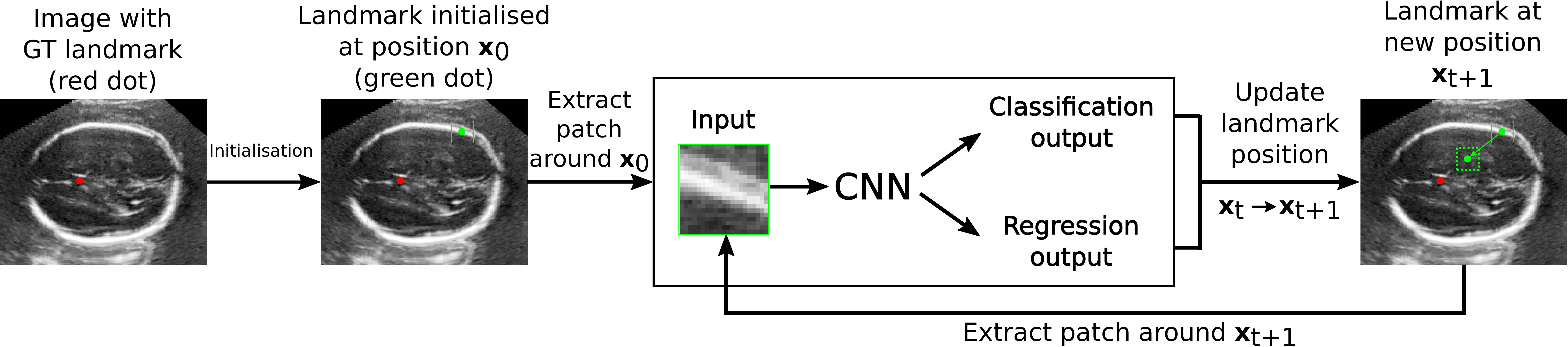}
\caption{Overall framework of PIN for single landmark localisation.}
\label{fig:pipeline}
\end{figure}
Recently, deep learning approaches have been proposed for this purpose \cite{Payer2016,Laina2017,Xu2017,Zheng2015,7961205,7493535} but there remain major challenges: (a) typically only a limited amount of annotated medical images is available, (b) model training and inference for 3D medical images is computationally intensive, making  real-time applications challenging and (c) when multiple landmarks are detected jointly, their spatial relationships should be taken into account.

\noindent \textbf{\textit{Related work:}} Deep learning methods for landmark localisation can be divided into two categories: The first category adopts an end-to-end learning strategy where the entire image is taken as input to a convolutional neural network (CNN) while the output is a map from which the landmark coordinates can be inferred directly. Payer \emph{et al.}~\cite{Payer2016} and Laina \emph{et al.}~\cite{Laina2017} output a heatmap in which Gaussians are located at the landmark positions. Xu \emph{et al.}~\cite{Xu2017} train a supervised action classifier (SAC) that outputs an action map whose classification labels denote the direction towards the true landmark location. However, end-to-end learning methods are typically applied to 2D images since 3D volumetric networks require large receptive fields for landmark tasks. Such 3D networks are computationally intensive, which inhibits real-time performance, and require a large amount of memory during training, which is beyond current hardware's capabilities. 

The second category uses image patches as training samples to learn a classification or regression model. Zheng \emph{et al.}~\cite{Zheng2015} extract a patch around each voxel in the image and use a neural network to classify if a landmark is present at the patch centre. Zhang \emph{et al.}~\cite{7961205} and Aubert \emph{et al.}~\cite{7493535} use a CNN-based regression model that learns the association between an image patch and its 3D displacement to the true landmark. Ghesu \emph{et al.}~\cite{Ghesu2016} propose a deep reinforcement learning (DRL) approach that also operates on patches. Most patch-based methods require dense sampling of many image patches during prediction which is computationally intensive. Furthermore, most methods require the training of separate models to detect each landmark. This is time-consuming and neglects the spatial relationships among multiple landmarks.

\noindent\textbf{\textit{Contribution:}} In this paper, we propose a novel landmark localisation approach that uses a patch-based CNN to predict multiple landmarks efficiently in an iterative manner. We term this approach Patch-based Iterative Network (PIN). PIN has distinct  advantages that address the key challenges of landmark localisation in 3D medical images: \textbf{(1)} During inference, PIN guides the patch towards the true landmark location using iterative sparse sampling. This approach reduces the computational cost by avoiding dense sampling at every voxel of the volume. \textbf{(2)} PIN uses a 2.5D representation to approximate the 3D patch as network input. This accelerates computation as only 2D convolutions are required. \textbf{(3)} PIN treats landmark localisation as a combined regression and classification problem for which a joint network is learned  via multi-task learning. This prevents model overfitting, improves generalisation ability of the learned features and increases localisation accuracy. \textbf{(4)} PIN detects multiple landmarks jointly using a single model and takes the global anatomical spatial relationships among landmarks into account. We evaluate  the landmark localisation accuracy of PIN using 3D ultrasound images of the fetal brain. In addition, clinically useful scan planes can be extracted from the predicted landmarks which visually resemble the anatomical standard planes as defined by fetal screening standards, \emph{e.g.}, \cite{screening2015}.

\section{Method}
\noindent\textbf{\textit{Overall Framework:}}
Fig.~\ref{fig:pipeline} illustrates the overall PIN framework for single landmark localisation. We show the 2D case for clarity but the method works similarly in 3D. Given an image, the goal is to predict the true landmark coordinates (red dot in Fig.~\ref{fig:pipeline}). A position $\boldsymbol{x}_0$ is first initialised at instant $t$=0 and a patch centred around $\boldsymbol{x}_0$ is extracted (solid green box in Fig.~\ref{fig:pipeline}). The CNN takes the patch as input and predicts regression and classification outputs that are used to compute a new position $\boldsymbol{x}_{t+1}$ from the previous position $\boldsymbol{x}_t$, bringing the patch closer to the true landmark location. The patch at $\boldsymbol{x}_{t+1}$ (dashed green box in Fig.~\ref{fig:pipeline}) is then given as input to the CNN and the process is repeated until the patch reaches the true landmark position.

\begin{figure}[htb]
\centering
\includegraphics[height=0.25\textheight]{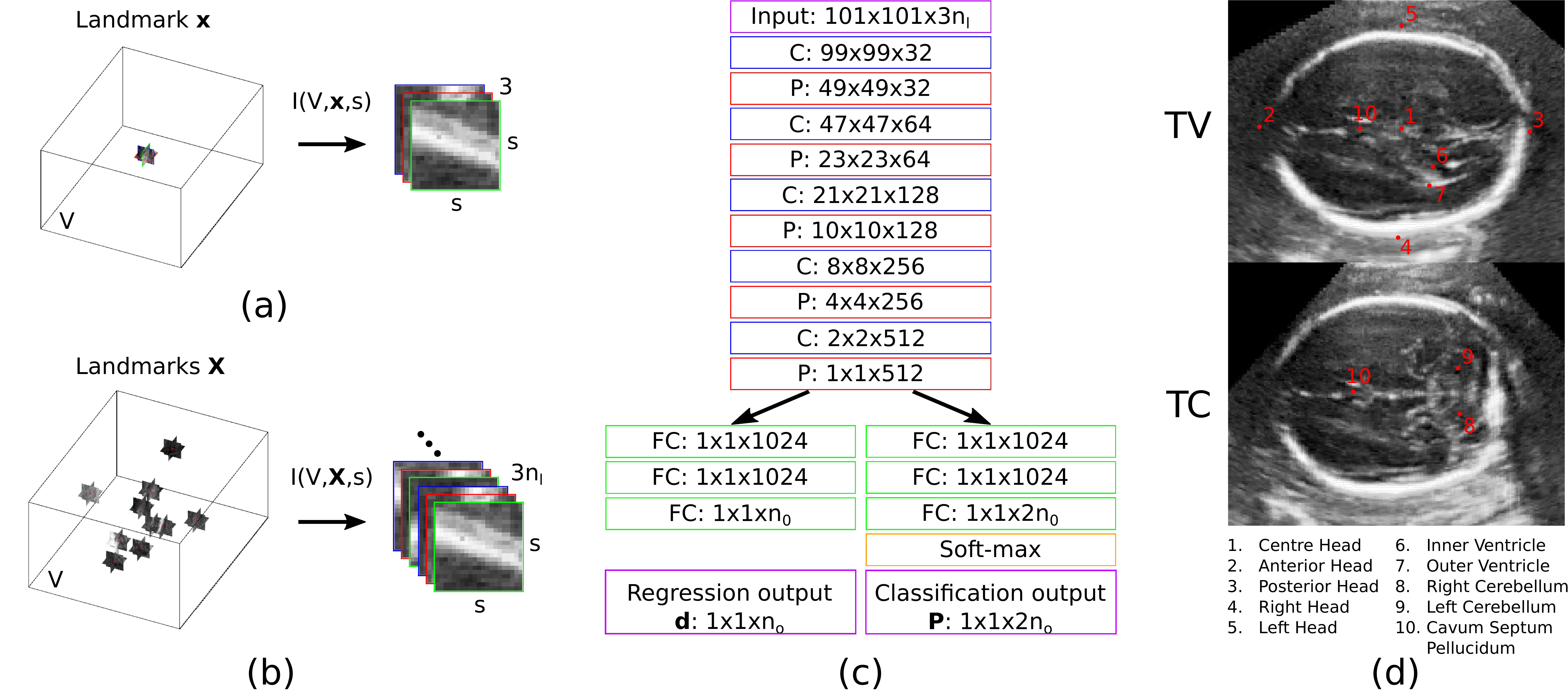}
\caption{(a) Patch extraction of a single landmark. (b) Patch extraction of multiple landmarks. (c) CNN architecture combining regression and classification. Output size of each layer is represented as $\textrm{width}\times \textrm{height}\times (\textrm{\# feature maps})$. (d) Landmarks defined on the TV and TC plane for fetal sonographic examination.}
\label{fig:architecture_patches_labels}
\end{figure}

\noindent\textbf{\textit{Network Input:}}
For 3D data, the CNN input can be a 3D volume patch. However, 3D convolution operations on volume patches are computationally expensive. To this end, we use a 2.5D representation to approximate the full 3D patch. Specifically, given a particular position $\boldsymbol{x}={(x,y,z)^{T}}$ in a volume $V$, we extract three 2D image patches centred around $\boldsymbol{x}$ at the three orthogonal planes (Fig.~\ref{fig:architecture_patches_labels}a). The patch extraction function is denoted as $I(V,\boldsymbol{x},s)$ where $s$ is the length of the square patch. The three 2D patches are then concatenated together as a 3-channel 2D patch which is passed as input to the CNN. Such a representation is computationally efficient since it requires only 2D convolutions and still provides a good approximation of the full 3D volume patch. 

\noindent\textbf{\textit{Joint Regression and Classification:}}
PIN jointly predicts the magnitude and direction of movement of a current point towards the true landmark by combining a regression and a classification task together in a multi-task learning framework. This joint framework shares model parameters in the convolutional layers and is experimentally shown to learn more generalisable features, which improves overall performance.

The regression task estimates how much the point at the current position should move to get to the true landmark location. The regression output $\boldsymbol{d}={(d_1,d_2,\hdots,d_{n_o})}^{T}$ is a displacement vector that predicts the relative distance between the current and true landmark positions. In single landmark localisation, $\boldsymbol{d}$ has $n_o=3$ elements which give the displacement along each coordinate axis. 

The classification task estimates the direction of current point movement towards the true landmark by dividing direction into 6 discrete classification categories: positive and negative direction along each coordinate axes~\cite{Xu2017}. Denoting $c$ as the classification label, we have $c\in \{ { c }_{ 1 }^{ + },{ c }_{ 1 }^{ - },{ c }_{ 2 }^{ + },{ c }_{ 2 }^{ - },{ c }_{ 3 }^{ + },{ c }_{ 3 }^{ - }\} $. For instance, ${ c }_{ 1 }^{+}$ is the category representing movement along the direction of positive x-axis. The classification output $\boldsymbol{P}$ is then a vector with $2n_o=6$ elements, each representing the probability/confidence of movement in that direction. Mathematically, $\boldsymbol{P}=(P_{ { c }_{ 1 }^{ + } },P_{ { c }_{ 1 }^{ - } },\hdots ,P_{ { c }_{ { n }_{ o } }^{ + } },P_{ { c }_{ { n }_{ o } }^{ - } })^{T}$ where $P_{ { c }_{ 1 }^{ + } }=\textrm{Prob}(c={ c }_{ 1 }^{ + })$.

Given a volume $V$ and its ground truth landmark point $\boldsymbol{x}^{GT}$, a training sample is represented by $(I(V,\boldsymbol{x},s), \boldsymbol{d}^{GT}, \boldsymbol{P}^{GT})$ where $\boldsymbol{x}$ is a point randomly sampled from $V$ and $I(V,\boldsymbol{x},s)$ is its associated patch. The ground truth displacement vector is given by $\boldsymbol{d}^{GT}=\boldsymbol{x}^{GT}-\boldsymbol{x}$. To obtain $\boldsymbol{P}^{GT}$, we first determine the ground truth classification label $c^{GT}$ by selecting the component of $\boldsymbol{d}^{GT}$ with the maximum absolute value and taking into account its sign,
\begin{equation}
c^{GT}=\begin{cases} {c}_{i}^{+},\qquad \textrm{if}\quad {d}_{i}^{GT}>0   \\ {c}_{i}^{-},\qquad \textrm{otherwise}. \end{cases}
\end{equation}
where $i=\textrm{argmax}(\textrm{abs}(\boldsymbol{d}^{GT}))$. For a vector $\boldsymbol{a}$, $\textrm{argmax}(\boldsymbol{a})$ returns the index of the vector component with maximum value. During training, a hard classification label is used. As such, the probability vector $\boldsymbol{P}^{GT}$ is obtained as a one-hot vector where component $P_{c^{GT}}$ is set to 1 and all others set to 0. The CNN is trained by minimising the following combined loss function:
\begin{equation} \label{eq:loss}
L = (1-\alpha )\frac { 1 }{ { n }_{ 0 }{ n }_{ batch } } \sum _{ n=1 }^{ n_{ batch } }{ { \left\| { \boldsymbol{d} }_{ n }^{ GT }-{ \boldsymbol{d} }_{ n } \right\|  }_{ 2 }^{ 2 } } -\alpha \frac { 1 }{ { n }_{ batch } } \sum _{ n=1 }^{ n_{ batch } }{ \log { ({ P }_{ { c }^{ GT },n } )}  } 
\end{equation}
\begin{figure}
\centering
\includegraphics[width=\linewidth]{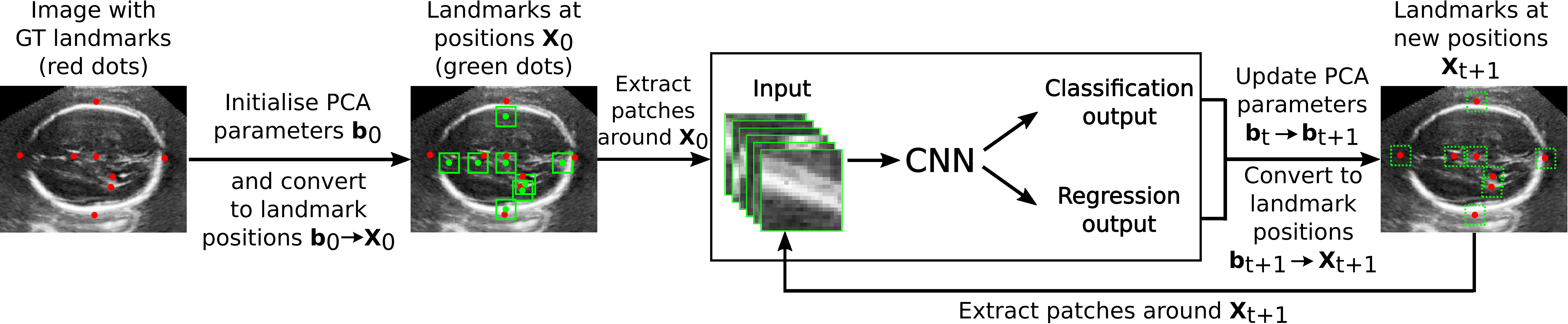}
\caption{Overall framework of PIN for multiple landmarks localisation.}
\label{fig:pipeline_multi}
\end{figure}
The first term is the Euclidean loss of the regression task and the second term is the cross-entropy loss of the classification task. $\alpha$ is the weighting between the two losses. $n_{batch}$ is the number of training samples in a mini-batch. ${ \boldsymbol{d} }_{ n }$ and ${ P }_{ { c }^{ GT },n }$ denote respectively the regression and classification outputs predicted by the CNN on the $n$th sample.

\noindent\textbf{\textit{CNN Architecture:}}
Fig.~\ref{fig:architecture_patches_labels}c shows the PIN CNN architecture combining the classification and regression tasks. The network comprises five convolution (C) layers, each followed by a max-pooling (P) layer. These layers are shared by both tasks. After the $5^{th}$ pooling layer, each task has three separate fully-connected (FC) layers to learn the task-specific features. All convolution layers use 3x3 kernels with stride=1 and all pooling layers use 2x2 kernels with stride=2. ReLU activation is applied after all convolution and FC layers except for the last FC layer of each task. Drop-out is added after each FC layer.

\noindent\textbf{\textit{PIN Inference:}}
Given an unseen 3D volume, we initialise 19 points in the volume (one at the volume centre and 18 others at fixed distance of one-quarter image size around it). The patch extracted from each point is forward passed into the CNN and the point is moved to its new position based on the CNN outputs ($\boldsymbol{d}$ and $\boldsymbol{P}$) and a chosen update rule. This process is repeated for $T$ iterations until there is no significant change in the displacement of the point. The final positions of the 19 points at iteration $T$ is averaged and taken to be the final landmark prediction. Multiple initialisations average out errors and improve the overall localisation accuracy. 

\noindent\textbf{\textit{PIN update rules:}} 
We proposed three update rules (A-C). Let $\boldsymbol{x}_{t}$ be the position of a point at iteration $t$ and $\boldsymbol{x}_{t+1}$ be the new updated position. Rule A is based only on the classification output $\boldsymbol{P}$. It updates the current landmark position by moving it one pixel in the direction category which has the highest probability as predicted by $\boldsymbol{P}$. Rule B is based only on the regression output $\boldsymbol{d}$ and is given by: $\boldsymbol{x}_{t+1}=\boldsymbol{x}_{t}+\boldsymbol{d}$. Rule C uses both the classification and regression outputs for the update and is given by: $\boldsymbol{x}_{t+1}=\boldsymbol{x}_{t}+{\boldsymbol{P}_{max}}\odot \boldsymbol{d}$ where $\odot$ is the element-wise multiplication operator and $\boldsymbol{P}_{max}=(\max{(P_{{ c }_{ 1 }^{ + }}, P_{{ c }_{ 1 }^{ - }})}, \max{(P_{{ c }_{ 2 }^{ + }}, P_{{ c }_{ 2 }^{ - }})}, \hdots , \max{(P_{{ c }_{ n_o }^{ + }}, P_{{ c }_{ n_o }^{ - }})})^{T}$. Intuitively, Rule C moves the point to its new position by an amount specified by the regression output weighted by a confidence probability specified by the classification output. This ensures smaller movement in the less confident direction and vice versa. 

\noindent\textbf{\textit{Multiple Landmarks Localisation:}}
The above approach for single landmark localisation has two drawbacks: \textbf{(1)} Separate CNN models are required for each landmark which increase the parametrisation significantly and thus computational cost for training and inference. \textbf{(2)} Individual landmark prediction ignores the anatomical relationships between the different landmarks. To overcome these problems, we extend our approach to localise multiple landmarks simultaneously using only one CNN model which also accounts for inter-landmark relationships by working in a reduced dimensional space.

Let $\boldsymbol{X}={(x_1,y_1,z_1,\hdots,x_{n_l},y_{n_l},z_{n_l})}^{T}$ be the 3D coordinates of all $n_l$ landmarks of one volume. Given a training set of $\boldsymbol{X}$, we use PCA to transform $\boldsymbol{X}$ into a lower dimensional space. The transformations between the original and reduced dimensional spaces are given by,
\begin{equation} \label{eq:b2x}
\boldsymbol{X}=\bar { \boldsymbol{X} } + \boldsymbol{W}\boldsymbol{b}
\end{equation}
\begin{equation} \label{eq:x2b}
\boldsymbol{b}={\boldsymbol{W}}^{T}(\boldsymbol{X}-\bar { \boldsymbol{X} }),
\end{equation}
where $\bar { \boldsymbol{X} }$ is the mean of the training set, $\boldsymbol{b}$ is a $n_b$-element vector where $n_{b}<3n_l$ and the columns of matrix $\boldsymbol{W}$ are the $n_{b}$ eigenvectors. In our case, $n_l=10$ and we set $n_b=15$ to explain 99.5\% of the total variations in the training set.

We can directly apply our PIN approach to the reduced dimensional space by replacing all occurrences of $\boldsymbol{x}$ by $\boldsymbol{b}$. Fig. \ref{fig:pipeline_multi} illustrates the PIN approach for multiple landmarks. Specifically, 3 orthogonal patches are extracted for every landmark and concatenated together so that a $s\times s\times 3n_l$ block is passed as CNN input (Fig. \ref{fig:architecture_patches_labels}b). $\boldsymbol{d}$ becomes the displacement vector in the reduced dimensional space with $n_o=n_b$ elements. The number of classification categories becomes $2n_b$ which include positive and negative directions along each dimension of the reduced space. Hence, $\boldsymbol{P}$ is a $2n_b$-element vector. Training can be carried out similar to Eq.~\ref{eq:loss} with the only difference being $\boldsymbol{d}^{GT}=\boldsymbol{b}^{GT}-\boldsymbol{b}$ where $\boldsymbol{b}^{GT}$ is transformed from $\boldsymbol{x}^{GT}$ using Eq.~\ref{eq:x2b} and $\boldsymbol{b}$ is randomly sampled. During inference, we update $\boldsymbol{b}$ iteratively using $\boldsymbol{b}_{t+1}=\boldsymbol{b}_{t}+{\boldsymbol{P}_{max}}\odot \boldsymbol{d}$ (Rule C) and use Eq.~\ref{eq:b2x} to convert $\boldsymbol{b}_{t+1}$ back to $\boldsymbol{X}_{t+1}$ for patch extraction in the next iteration. We use multiple initialisations of $\boldsymbol{b}_0$ (one initialisation with $\boldsymbol{b}_0=\boldsymbol{0}$ and five random initialisations) and take their mean results as the final landmarks prediction.

\section{Experiments and Results}
\noindent\textbf{\textit{Data:}}
PIN is evaluated on 3D ultrasound volumes of the fetal head from 72 subjects. Each volume is annotated by a clinical expert with 10 anatomical landmarks that lie on two standard planes (transventricular (TV) and transcerebellar (TC)) commonly used for fetal sonographic examination as defined in the UK FASP handbook~\cite{screening2015} (Fig. \ref{fig:architecture_patches_labels}d). 70\% of the dataset is randomly selected for training and the remaining 30\% is used for testing. All volumes are processed to be isotropic and resized to 324$\times$207$\times$279 voxels with voxel size 0.5$\times$0.5$\times$0.5 mm\textsuperscript{3}.

\noindent\textbf{\textit{Experiment Setup:}}
PIN is implemented using Tensorflow running on a machine with Intel Xeon CPU E5-1630 at 3.70 GHz and one NVIDIA Titan Xp 12GB GPU. Patch size $s$ is set to 101. During training, we set $n_{batch}$=64. Weights are initialised randomly from a distribution with zero mean and 0.1 standard deviation. Optimisation is carried out for 100,000 iterations using the Adam algorithm with learning rate=0.001, $\beta_1$=0.9 and $\beta_2$=0.999. We choose $\alpha$=0.5 empirically unless otherwise stated. During inference, $T$=350 for Rule A and $T$=10 for Rule B and C.

\begin{table}[htb]
\centering
\caption{Localisation error (mm) and runtime (s) of different approaches for single landmark (CSP) localisation. C and R denote classification and regression training loss respectively. Results presented as (Mean $\pm$ Standard Deviation).}
\label{table:ResultSingleLandmark}
\resizebox{0.8\textwidth}{!}{%
\begin{tabular}{|l|c|c|c|c|c|c|c|}
\hline
 & PIN1 & PIN2 & PIN3 & PIN4 & PIN5 & DRL \cite{Ghesu2016} \\ \hline
Training loss & C & R & C+R & C+R & C+R & - \\ \hline
Inference rule & Rule A & Rule B & Rule A & Rule B & Rule C & - \\ \hline
Localisation error & 7.53$\pm$6.48 & 6.45$\pm$3.96 & 6.34$\pm$3.62 & 6.08$\pm$3.90 & \textbf{5.47$\boldsymbol{\pm}$4.23} & 7.37$\pm$5.86 \\ \hline
Running time (s) & 3.56 & 0.09 & 3.50 & 0.09 & \textbf{0.09} & 6.58 \\ \hline
\end{tabular}}
\vspace{-0.5cm}
\end{table}

\begin{table}[]
\centering
\caption{Localisation error (mm) of PIN for single and multiple landmark localisation. Results presented as (Mean $\pm$ Standard Deviation).}
\label{table:ResultMultipleLandmarks}
\resizebox{\textwidth}{!}{%
\begin{tabular}{|l|c|c|c|c|c|c|c|c|c|c|c|}
\hline
Landmarks & 1 & 2 & 3 & 4 & 5 & 6 & 7 & 8 & 9 & 10 & Overall \\ \hline
PIN-Single & 5.62$\pm$2.85 & 11.30$\pm$7.24 & 8.13$\pm$3.90 & 7.23$\pm$3.73 & 7.11$\pm$4.73 & 4.39$\pm$2.07 & 5.45$\pm$2.73 & 4.04$\pm$2.22 & 5.50$\pm$3.64 & 5.47$\pm$4.23 & 6.42$\pm$4.49 \\ \hline
PIN-Multiple & 4.34$\pm$2.21 & 8.80$\pm$4.27 & 6.28$\pm$2.77 & 6.31$\pm$3.32 & 5.56$\pm$2.71 & 4.68$\pm$2.27 & 5.15$\pm$2.90 & 4.70$\pm$2.33 & 4.57$\pm$1.92 & 5.50$\pm$2.79 & \textbf{5.59$\boldsymbol{\pm}$3.09} \\ \hline
\end{tabular}}
\vspace{-0.5cm}
\end{table}

\noindent\textbf{\textit{Results:}}
Table~\ref{table:ResultSingleLandmark} compares the landmark localisation errors of a single landmark, cavum septum pellucidum (CSP), using several PIN variants which differ in the CNN model training and the inference update rule. Given the same inference rules, the model trained using both classification and regression losses ($\alpha$=0.5) achieves lower error than the models trained using either loss alone ($\alpha$=1 or 0) (PIN1 vs PIN3, PIN2 vs PIN4). This illustrates the benefits of multi-task learning. Using the model trained with joint losses, we then compare the effect of different inference rules. PIN3 uses only the classification output which can result in landmarks getting stuck and oscillating between two opposing classification categories during iterative testing (\emph{e.g.}, $c_{1}^{+}$ and $c_{1}^{-}$). PIN3 also takes longer during inference since the patch moves by one pixel at each test iteration and requires more iterations to converge. PIN4 uses only the regression output, which improves the localisation accuracy and runtime as the patch `jumps' towards the true landmark position at each iteration. This requires much fewer iterations to converge. PIN5 achieves the best localisation accuracy by combining the classification and regression outputs where the regression output gives the magnitude of movement weighted by the classification output giving the probability of movement in each direction. Our proposed PIN approach also outperforms a recent state-of-the-art landmark localisation approach using DRL~\cite{Ghesu2016}.

Table~\ref{table:ResultMultipleLandmarks} shows the localisation errors for all ten landmarks. PIN-Single trains a separate model for each landmark while PIN-Multiple trains one joint model that predicts all the landmarks simultaneously. Since PIN-Multiple accounts for anatomical relationships among the landmarks, it has a lower overall localisation error than PIN-Single. PIN-Single needs a total of 0.94s to predict all ten landmarks in sequence while PIN-Multiple needs 0.44s to predict all ten landmarks simultaneously. 
\begin{figure}[htb]
\centering
\includegraphics[width=\linewidth]{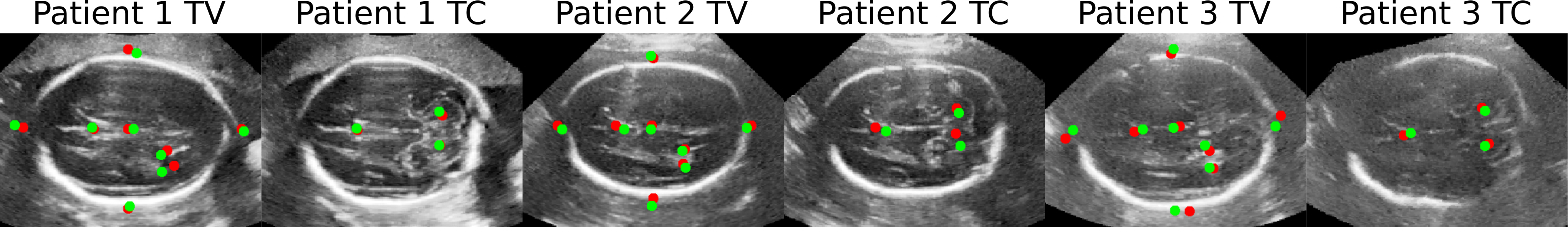}
\caption{Visualisation of landmarks predicted by PIN-Multiple (green dots) vs. ground truth landmarks (red dots).}
\label{fig:results}
\end{figure}Fig.~\ref{fig:results} shows the TV and TC planes containing the ground truth landmarks as red dots. The landmarks predicted by PIN-Multiple are projected onto these standard planes as green dots. The supplementary materials provide visual comparison of standard planes obtained from ground truth and predicted landmarks as well as videos showing several initialisations converging towards the true landmark positions (and standard planes) after ten inference updates.

\section{Conclusion}
We have presented PIN, a new approach for anatomical landmark localisation. Its patch-based and iterative nature enables training on limited data and fast prediction on large 3D volumes. A joint regression and classification model is trained by multi-task learning to improve localisation accuracy. PIN is capable of multiple landmark localisation and uses PCA to impose anatomical constraints among landmarks. PIN is generic to landmark localisation and as future work, we are extending PIN to other medical applications. It is also worthwhile to replace PCA with an autoencoder to model non-linear correlations among landmarks.

\subsubsection*{Acknowledgments.} Supported by the Wellcome Trust IEH Award [102431]. The authors thank Nvidia Corporation for the donation of a Titan Xp GPU. 
\bibliography{References}

\begin{thebibliography}{1}
\providecommand{\url}[1]{\texttt{#1}}
\providecommand{\urlprefix}{URL }

\bibitem{7493535}
Aubert, B., Vazquez, C., Cresson, T., Parent, S., Guise, J.D.: Automatic spine
  and pelvis detection in frontal x-rays using deep neural networks for patch
  displacement learning. In: ISBI 2016. pp. 1426--1429 (April 2016)

\bibitem{Ghesu2016}
Ghesu, F.C., Georgescu, B., Mansi, T., Neumann, D., Hornegger, J., Comaniciu,
  D.: An artificial agent for anatomical landmark detection in medical images.
  In: MICCAI 2016. pp. 229--237 (2016)

\bibitem{Laina2017}
Laina, I., Rieke, N., Rupprecht, C., Vizca{\'i}no, J.P., Eslami, A., Tombari,
  F., Navab, N.: Concurrent segmentation and localization for tracking of
  surgical instruments. In: MICCAI 2017. pp. 664--672 (2017)

\bibitem{screening2015}
{NHS}: Fetal anomaly screening programme: programme handbook June 2015. Public
  Health England (2015)

\bibitem{Payer2016}
Payer, C., {\v{S}}tern, D., Bischof, H., Urschler, M.: Regressing heatmaps for
  multiple landmark localization using cnns. In: MICCAI 2016. pp. 230--238
  (2016)

\bibitem{Xu2017}
Xu, Z., Huang, Q., Park, J., Chen, M., Xu, D., Yang, D., Liu, D., Zhou, S.K.:
  Supervised action classifier: Approaching landmark detection as image
  partitioning. In: MICCAI 2017. pp. 338--346 (2017)

\bibitem{7961205}
Zhang, J., Liu, M., Shen, D.: Detecting anatomical landmarks from limited
  medical imaging data using two-stage task-oriented deep neural networks. IEEE
  Transactions on Image Processing  26(10),  4753--4764 (Oct 2017)

\bibitem{Zheng2015}
Zheng, Y., Liu, D., Georgescu, B., Nguyen, H., Comaniciu, D.: 3d deep learning
  for efficient and robust landmark detection in volumetric data. In: MICCAI
  2015. pp. 565--572 (2015)

\end{thebibliography}
\end{document}